\documentclass[11pt]{article} 
\usepackage{acl}
\usepackage{xcolor}
\usepackage{hyperref}
\usepackage{url}
\usepackage{verbatim}
\usepackage{multirow}
\usepackage{enumitem}
\setlist[itemize]{leftmargin=0.5cm}

\usepackage{latexsym}
\usepackage{booktabs} 
\usepackage{makecell}

\usepackage{subfig}

\usepackage[T1]{fontenc}
\usepackage[utf8]{inputenc}

\usepackage{microtype}

\usepackage[cmex10]{amsmath}
\usepackage{amsfonts, amsthm, amssymb}
\usepackage{pifont}

\definecolor{olivegreen}{RGB}{85, 107, 47}

\usepackage{algorithm}
\usepackage[noend]{algpseudocode}
\usepackage{caption}

\usepackage{tikz}
\usetikzlibrary{arrows, decorations.pathreplacing, decorations.pathmorphing,backgrounds,positioning,fit,petri}

\usepackage[retainorgcmds]{IEEEtrantools}
\newcommand{\be}{\begin{IEEEeqnarray*}{rCl}}
\newcommand{\ee}{\end{IEEEeqnarray*}}

\newcommand{\ben}{\begin{IEEEeqnarray}{rCl}}
\newcommand{\een}{\end{IEEEeqnarray}}

\usepackage{tcolorbox}
\newtcbox{\mybox}[1][blue]{on line,
arc=0pt,outer arc=0pt,colback=#1!10!white,colframe=#1!50!black,
boxsep=0pt,left=2pt,right=2pt,top=2pt,bottom=2pt,
boxrule=0pt,bottomrule=0pt,toprule=0pt}

\title{Skywork: A More Open Bilingual Foundation Model}
\author{\small {Tianwen Wei, Liang Zhao, Lichang Zhang, Bo Zhu, Lijie Wang, Haihua Yang, Biye Li, Cheng Cheng,  Weiwei L\"u, Rui Hu} \\
\small {Chenxia Li, Liu Yang, Xilin Luo, Xuejie Wu, Lunan Liu, Wenjun Cheng, Peng Cheng, Jianhao Zhang, Xiaoyu Zhang} \\
\small {Lei Lin, Xiaokun Wang, Yutuan Ma, Chuanhai Dong, Yanqi Sun, Yifu Chen, Yongyi Peng, Xiaojuan Liang} \\
\small {Shuicheng Yan, Han Fang, Yahui Zhou}\thanks{\quad Email: \texttt{\{forename\}.\{surname\}@kunlun-inc.com}} \\
\quad \\
Skywork Team, Kunlun Inc.
}

\begin{document}
\maketitle
\begin{abstract}
In this technical report, we present Skywork-13B, a family of large language models (LLMs) trained on a corpus of over 3.2 trillion tokens drawn from both English and Chinese texts. This bilingual foundation model is the most extensively trained and openly published LLMs of comparable size to date. We introduce a two-stage training methodology using a segmented corpus, targeting general purpose training and then domain-specific enhancement training, respectively. We show that our model not only excels on popular benchmarks, but also achieves \emph{state of the art} performance in Chinese language modeling on diverse domains. 
Furthermore, we propose a novel leakage detection method, demonstrating that data contamination is a pressing issue warranting further investigation by the LLM community.
To spur future research, we release Skywork-13B along with checkpoints obtained during intermediate stages of the training process. We are also releasing part of our SkyPile corpus, a collection of over 150 billion tokens of web text, which is the largest high quality open Chinese pre-training corpus to date. 
We hope Skywork-13B and our open corpus will serve as a valuable open-source resource to democratize access to high-quality LLMs.
\end{abstract}

\section{Introduction}
Natural Language Processing (NLP), a vital branch of artificial intelligence, has experienced a transformative surge in recent years. Pivotal to this revolution has been the advent and advancement of large language models (LLMs) \cite{instruct_gpt, gpt4_report, gpt4_sparks, palm, palm2, llama, llama2}. These complex computational structures, composed of billions of parameters, are capable of understanding, generating, and translating human language with an unprecedented degree of accuracy and sophistication. However, the proliferation of these models has also been accompanied by a growing trend towards commercialization and a lack of transparency, a phenomenon that is increasingly influencing the dynamics of the open-source community.

Historically, the open-source community has thrived on the principles of collaboration, transparency, and unrestricted sharing of ideas. However, as the commercial potential of LLMs has been recognized, this openness has begun to diminish. The reality is that many organizations only make model checkpoints publicly accessible, while withholding vital information on model reproduction. This practice significantly hampers the progress of the field.

In an effort to revive the spirit of the open-source community and contribute to the ongoing dialogue about transparency in AI, we present Skywork-13B: a family of bilingual large language models with 13 billion parameters, trained on a colossal corpus of more than 3.2 trillion tokens drawn from both English and Chinese texts. To our knowledge, our Skywork-13B is the most thoroughly trained family of open LLMs of comparable size to date.

In this technical report, we offer a comprehensive disclosure of the Skywork-13B developmental journey. We detail the composition of our training data, provide insights into the evolutionary trajectory of the model’s abilities during training, and share methodologies that could be employed to enhance model ability in specific domains. We believe that such an open approach not only aids in the reproducibility of our work but also provides a valuable resource for other researchers seeking to explore and expand the capabilities of large language models.
This technical report is also a call to action for renewed transparency in the field of NLP. Through it, we hope to inspire a return to a more collaborative, open-source community, where progress is not hampered by commercial considerations but propelled by collective intelligence and shared wisdom.

Our contributions are the following:
\begin{itemize}
\item We release Skywork-13B\footnote{Github repository: {\url{https://github.com/SkyworkAI/Skywork}}.}, a family of LLMs that is the most extensively trained and openly published LLMs of comparable size to date. Our Skywork-13B family includes 1) Skywork-13B-Base, a strong foundation model with \emph{state of the art} Chinese language modeling capability, and 2) Skywork-13B-Chat, a fined-tuned version optimized for conversation\footnote{In this technical report we focus on the development of the base model. Details on Skywork-13B-Chat can be found in our Github repository.}.
\item We disclose detailed information on the training process and data composition. We also release intermediate checkpoints, which provide a valuable resource for understanding how the model's capabilities develop over the course of training. It enables other researchers to leverage these checkpoints for their specific use-cases.
\item We release a portion of our high quality training corpus, totaling more than 150 billion tokens. To our knowledge, this is the largest open Chinese corpus for language model pre-training to date. 
\item We develop a novel method that detects the level of in-domain data usage during the training stage. To facilitate reproduction of the experiments presented in this report, we have released the relevant data.
\end{itemize}

\section{Methodology}
\subsection{Two Pre-training Stages \label{two_stages}}
In order to train Skywork-13B, we constructed SkyPile (see Section \ref{skypile}), a massive training corpus primarily constituted by publicly accessible web pages. We identified a small subset of SkyPile, encompassing exercises and solutions that span a broad spectrum of subjects from primary to graduate school. This includes coding problems, national exam questions, textbook exercises, and others. Given the majority of these exercises are STEM-related, we henceforth refer to this subset and its complement as SkyPile-STEM and SkyPile-Main, respectively.

Rather than training the Skywork-13B foundation model directly on SkyPile as a whole, we adopted a two-stage training approach. The first stage, which constitutes the primary pre-training phase, involves training the model from scratch on SkyPile-Main. In the second stage, our Skywork-13B is enriched with STEM-related domain knowledge and problem-solving skills through continual pre-training on SkyPile-STEM. To circumvent the potential issue of catastrophic forgetting, this continual pre-training is performed on a mix of SkyPile-STEM and SkyPile-Main, rather than exclusively on SkyPile-STEM.

The decision to segregate Stage-1 and Stage-2 pre-training serves a dual purpose. Firstly, we acknowledge that a significant proportion of the samples from SkyPile-STEM are, by their nature, supervised data. Those data are closely related to popular benchmarks such as CEVAL \cite{ceval}, MMLU \cite{mmlu} and GSM8K \cite{gsm8k}, and can be utilized in a supervised fine-tuning (SFT) process to directly enhance model performance on related downstream tasks. In this context, the separation between Stage-1 and Stage-2 training enables us to more effectively assess the impacts of general-purpose pre-training (on web texts) and targeted pre-training (on in-domain/supervised data). Such insights could inform future data collection and compilation strategies for foundational model training.

Secondly, by restricting first stage pre-training to general-purpose data, we are able to produce a version of foundation model as an alternative to the one with targeted enhancement. While the latter demonstrates superior performance on certain downstream tasks, it is less capable in language modeling of natural texts. We posit that this alternative is a valuable contribution to the community, given its potential to excel in applications that do not require STEM-related competencies.

\subsection{Training Progress Monitoring }
It is of vital importance to monitor and assess progress made during pre-training in real-time.
Existing methods such as monitoring training loss and benchmark results on intermediate checkpoints, however, have their limitations.

The main issue of monitoring training loss lies in that its effectiveness comes into question when considering the potential of overfitting. The training loss is equivalent to validation loss only if the training data is utilized exactly once (i.e., in one epoch). Yet, in practical scenarios of training LLMs, high-quality data often go through the training process multiple times \cite{galactica, llama, code_llama, phi1, phi2}. Besides, even after explicit de-duplication, there may still exist significant amount of duplicated data in the training set \cite{slimpajama, semdedup}. 
In either cases, solely relying on training loss can lead to overlooking the issue of overfitting, thereby producing overly optimistic estimates of model performance.
The top left subplot in Figure \ref{validation_loss} illustrates the trajectory of the pre-training loss for our Skywork-13B model. Consistent with findings reported in \cite{llama,llama2,baichuan}, the loss demonstrates a steady decline throughout the training process. However, an observation not disclosed in these cited works is the behavior of the validation loss on held-out sets. From the figure it can be clearly seen that the validation losses seem to level off as training approaches its final stages.

Benchmarking based on intermediate checkpoints is another common monitoring approach \cite{llama, baichuan}. Nevertheless, it presents several challenges. Firstly, there is a high variance in benchmark results, which can lead to unstable and unreliable assessments of training progress. Secondly, benchmark results are not sensitive to minor progress in training. This insensitivity makes it difficult to accurately track gradual improvements during the training process. Besides, weaker models do not follow instructions well. Hence benchmark results may not accurately reflect their true learning progress or potential.   Finally, an inconvenience posed by most benchmarks is the necessity for model generation. This process is notably resource-intensive, demanding substantial computational power.

\begin{figure}[t]
\centering
\includegraphics[width=0.4\textwidth]{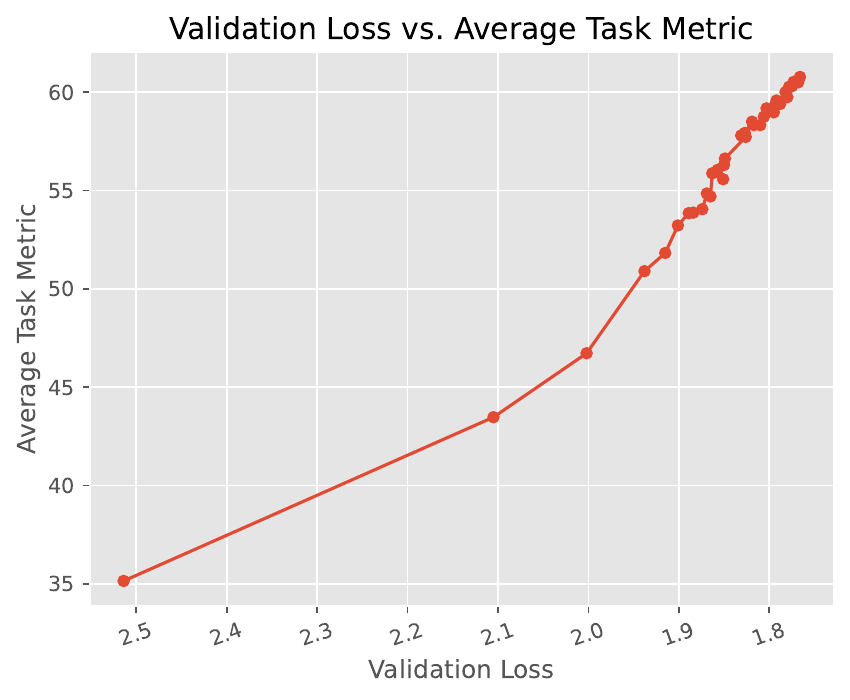}
\caption{Validation loss on English web texts vs. average task metric during the pre-training of Skywork-13B. The tasks include BoolQ \cite{boolq}, PIQA \cite{piqa}, Winogrande \cite{winogrande}, TriviaQA \cite{triviaqa} and RACE \cite{race}. 
}
\label{fig:loss_vs_metric}
\end{figure}

During the pre-training of Skywork-13B, we embrace the method of monitoring the language modeling loss across numerous reserved validation sets, each reflecting a distinct data distribution. More specifically, we have created separate validation sets for code, academic publications, social media posts, web texts in Chinese and English, among others. Conventional monitoring metrics are also utilized, but they serve merely as supplementary tools. In Figure \ref{fig:loss_vs_metric} we plot the curve of language model validation loss on English web texts against the average metric of several English downstream tasks. It is apparent that there is a very high correlation between the two quantities, showing that validation loss can serve as a valid proxy metric for downstream task performance.
 In the context of LLM pre-training, this approach also yields several other benefits:
\begin{itemize}
\item Ease of construction: Crafting multiple validation sets is a relatively effortless task. This enables the evaluation of a model's language modeling performance across varied domains.
\item Simplicity in computation: Calculation of validation loss is straightforward, significantly reducing the computational and logistical overhead associated with tracking model training.
\item High sensitivity to training progress: Validation loss is finely attuned to the progression of training, thereby offering a more detailed perspective on how models evolve and improve over time.
\item Model-agnosticism: Validation loss is indifferent to the composition of the training corpus or the model architecture. It allows for comparison not only between different checkpoints produced within a single training session, but also across varied models from the community. This ensures a consistent and equitable basis for model comparison.
\end{itemize}
Note that monitoring the validation loss on a held-out set sharing the same distribution as the training set is a ubiquitous practice in machine learning. However, the observation of validation loss across multiple held-out sets, each with deliberate, unique distributions, is not common.
We also note that the perspective asserting the primacy of language modeling loss as the paramount performance metric for models is not a recent revelation. This principle has been either explicitly or implicitly adopted in a number of research studies, as exemplified in \cite{kaplan2020scaling, hoffmann2022scaling, palm2, xia2023, deletang2023}.

\section{Pre-training}
\subsection{SkyPile Corpus \label{skypile}}
In order to train Skywork-13B, we build SkyPile, a vast, high quality corpus comprising more than 6 trillion tokens.  A segment of the corpus, comprising over 150 billion tokens of web text, has been open sourced to facilitate research and training on Chinese LLMs\footnote{\url{huggingface.co/datasets/Skywork/SkyPile-150B}}.

Our SkyPile is an amalgamation of several sources, the overwhelming majority of which is gleaned from publicly accessible channels. 
Numerous prior research works, exemplified by initiatives such as LLaMA \cite{llama} and RefinedWeb \cite{refinedweb}, have substantiated the notion that publicly accessible web data can yield exceptionally high-quality LLMs. In alignment with this empirical evidence, we subscribe to the premise of leveraging publicly accessible webpages as our primary source for training data.

The construction of SkyPile is characterized by a dedicated emphasis on two primary dimensions: text quality and information distribution. Our data processing pipeline, inspired by \cite{ccnet, llama, refinedweb}, incorporates the following stages:

\begin{itemize}
\item {\bf Structural Extraction: } 
Due to the predominant source of our dataset being publicly accessible web pages, the objective of the first stage is the extraction of pertinent content while concurrently expunging extraneous textual elements that are deemed non-contributory to the training of our language model, e.g. these superfluous components include navigational bars, site-specific contact information, disjunctive title texts devoid of substantive content, etc. Subsequent to this culling process, the retained information predominantly consists of contiguous, medium to long-form textual passages. 

\item {\bf Distribution Filtering: }
In the pursuit of cultivating a profoundly adept LLM, the model's exposure must encompass a diverse array of content spanning an extensive spectrum of domains. Prior endeavors within the field have entailed the task of assigning categorical labels to each individual document or webpage, thereby manually dictating the composition of the training corpus. However, we posit that the corpus employed for LLM training has burgeoned to such an extent that the knowledge it encapsulates can not be compartmentalized discretely. Consequently, eschewing a label-centric approach, our methodology centers on benchmarking the semantic affinities existing between textual segments, thereby identifying and omitting those text blocks characterized by an exceedingly high recurrence rate.

\item {\bf Deduplication: } 
Deduplication has demonstrated its remarkable efficacy in enhancing the overall quality of a training corpus, and it has found extensive application in virtually all prominent datasets \cite{hernandez2022scaling, kandpal2022, semdedup, lee2022dedup}. Within the framework of SkyPile, we regard deduplication as an integral component of the Distribution Filtering process. When considering the broader perspective, it becomes evident that duplication constitutes a paramount factor influencing the semantic distribution of a corpus. Consequently, the techniques and strategies we employed during the distribution filtering phase autonomously eliminated a substantial portion of duplicated content.

\item {\bf Quality Filtering: }
In this phase, we deploy the CCNet \cite{ccnet} pipeline to perform two critical filtration tasks: the elimination of content of inferior quality and the exclusion of pages that are neither in English nor Chinese. We trained a binary classifier that predicts the likelihood that a given webpage is suitable for inclusion as a reference within the Wikipedia corpus. The outcome of this stage is organized into distinct quality-based categories, and we  retain exclusively the high quality groups, opting to discard the remaining groups in its entirety.
\end{itemize}
Above we described our pre-processing pipeline for natural text. As for Github content, 
we employ an approach that is similar to \cite{redpajama}. We have devised a collection of straightforward yet efficacious heuristics, encompassing criteria such as line length filtration and alphanumeric thresholds, designed to discern and exclude content of low quality. Our criteria are specifically oriented toward enhancing content quality, as opposed to merely curbing its volume. Notably, in contrast to prevailing practices that involve the wholesale removal of a significant portion of json, xml, yaml, and html content, we have made a deliberate choice to retain a judiciously proportionate representation of these data formats.

Note that in pursuit of harmonizing the model's proficiency in both English and Chinese, we include in SkyPile a curated high-quality parallel corpora. This data is meticulously structured to pair a complete English paragraph with its corresponding Chinese counterpart, ensuring a seamless alignment of linguistic capabilities between the two languages.

\subsection{Training Data Composition}

Our Skywork-13B is pre-trained for 3.2 trillion tokens, sampled from SkyPile. 
Texts from certain sources are deemed as of high quality, e.g. Wikipedia, hence have undergone upsampling. However, we generally stick to the rule that the number of repetition does not exceed five, as is recommended by recent studies \cite{galactica, muennighoff2023}. 

We report in Table \ref{training_corpus} a breakdown of the constituent components of the training tokens during Stage-1 pre-training. 
The training tokens are primarily composed of English and Chinese texts, constituting 49.8\% and 39.6\% of the data, respectively. Code contributes 8.0\% to the total, with texts in other languages accounting for the remaining 2.4\%. The category labeled as ``miscellany'' encompasses a diverse range of texts, including but not limited to, legal articles, court documents, company annual reports, and classical literature.

\begin{table}[t]
\renewcommand{\arraystretch}{1.2} 
\centering
\resizebox{0.45\textwidth}{!}{%
\begin{tabular}{c|l|c}
\toprule
 & \textbf{Category} &  \textbf{Percentage} \\ 
 \midrule
\multirow{5}{*}{\textbf{English}} 
& Webpages &  39.8\% \\ 
& Books &  3.6\% \\ 
& Academic Papers &  3.0\% \\
& Encyclopedia &  0.5\% \\ 
& Miscellany   &  2.9\% \\ 
\midrule
\multirow{5}{*}{\textbf{Chinese}} 
& Webpages &  30.4\% \\ 
& Social Media & 5.5\% \\
& Encyclopedia &   0.8\%  \\
& Miscellany  &  3.1\%  \\
\midrule
\textbf{Other Lang.} & Encyclopedia & 2.4\% \\
\midrule
\textbf{Code} & Github & 8.0\% \\
\bottomrule
\end{tabular}
}
\caption{Breakdown of training data in Stage-1 pre-training of Skywork-13B.}
\label{training_corpus}
\end{table}

\subsection{Tokenizer}
We tokenize the data using byte-pair encoding (BPE) as implemented in SentencePiece 
\cite{sentencepiece2018}, following the approach of LLaMA \cite{llama}. Since our model is intended to be English-Chinese bilingual, we extend the original vocabulary of LLaMA, which primarily consists of latin-based words and subwords, with frequently used Chinese characters and words. Specifically, we add 8000 single-character tokens from BERT's vocabulary \cite{BERT} to LLaMA's vocabulary. We further expand the vocabulary with 25k frequent Chinese multi-character words. This results in a total vocabulary size of 65,536 tokens, of which 17 are reserved as special symbols.

As in LLaMA, we split all numbers into individual digits, and fall back to bytes to decompose unknown UTF-8 characters.

\begin{table}[ht]
\renewcommand{\arraystretch}{1.0} 
\centering
\resizebox{0.45\textwidth}{!}{%
\begin{tabular}{rl}
\toprule
Category & Size \\
\midrule
Latin based words \& subwords & 32000 \\
Chinese characters \& Unicode symbols& 8000  \\
Chinese words & 25519   \\
Reserved symbols & 17   \\
\midrule
\textbf{Total} & \textbf{65536}   \\
\bottomrule
\end{tabular}
}
\caption{Breakdown of the vocabulary used in Skywork-13B.}
\label{vocabulary}
\end{table}

\subsection{Architecture}
\begin{figure}[h]
\centering
\includegraphics[width=0.45\textwidth]{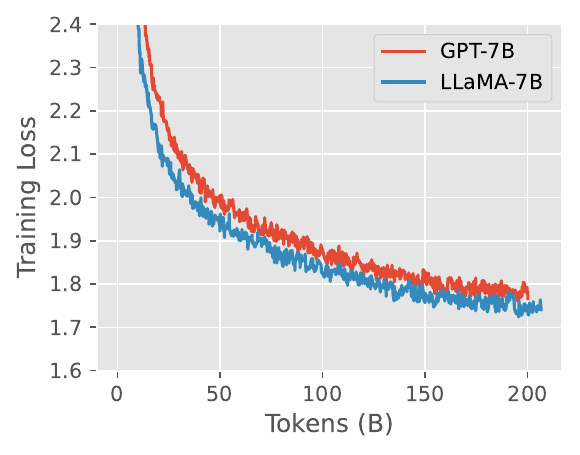} 
\caption{Preliminary Experiments: Comparison of conventional GPT architecture and more recent LLaMA architecture. For each of the two transformer variants, a model with 7 billion parameters is trained from Scratch on 200 Billion Tokens. The plot clearly shows that the LLaMA architecture achieves a lower training loss than GPT, demonstrating the former's superiority.}
\label{gpt_vs_llama}
\end{figure}

Our Skywork-13B is based on the transformer architecture \cite{VASWANI}, consisting of stacks of transformer-decoder layers. 
In contrast to the original transformer model, we have incorporated several modifications, inspired by LLaMA \cite{llama, llama2}. Our preliminary experiments, as illustrated in Figure \ref{gpt_vs_llama}, validate these changes, demonstrating the improved performance they confer. 
Details on this experiment can be found in Appendix \ref{GPT_vs_LLAMA}.

While our network architecture takes after the LLaMA model to a great extent, there exists a notable difference in our preference for a deeper, yet narrower, network. A comparative exploration of the Skywork-13B and LLaMA2-13B network configurations is presented in Table \ref{table:architecture}.

The specific modifications made are described in detail below.

\begin{itemize}
\item {\bf Positional Embedding:}
We use Rotary Positional Embedding (RoPE) \cite{rope}, that was motivated by its extensive adoption in various prominent large language models, such as LLaMA and PaLM, as well as its demonstrated effectiveness in extending the length of context windows, as evidenced by recent studies \cite{positional_interpolation, code_llama, effective_scaling}.
\item {\bf Layer Normalization:}
We replaced the conventional layer normalization with RMSNorm \cite{rmsnorm}. Additionally, we adopted pre-normalization in each layer instead of post-normalization, which has been shown to enhance the training stability of transformer models.
\item {\bf Activation:}
We employed the SwiGLU activation function \citep{swiglu}. In line with established conventions in prior studies, we reduced the dimension of the feed-forward network (FFN) from four times the hidden size to eight-thirds of the hidden size. This adjustment was made to maintain parity between the total parameters in a layer and those in the vanilla transformer layer.
\end{itemize}

\begin{table}[ht]
\renewcommand{\arraystretch}{1.0} 
\centering
\resizebox{0.45\textwidth}{!}{%
\begin{tabular}{r|cc}
\toprule
 & LLaMA2-13B & Skywork-13B \\
\midrule
Vocab. Size  & 32,000 & 65,536 \\
Hidden Dim. & 5,120 & 4,608\\
FFN Dim.    & 13,696 & 12,288\\
Head Dim.   & 128 & 128 \\
Num. Heads  & 40 & 36\\
Num. Layers & 40 & 52 \\
\midrule 
Seq. Len. & 4,096 & 4,096 \\
\#Tokens per Batch & 4M & 16M \\
Peak LR & 3e-4 & 6e-4 \\
Minimum LR & 3e-5 & 6e-5 \\
\bottomrule
\end{tabular}
}
\caption{Comparisons in architecture and important hyper-parameters of Skywork-13B and LLaMA2-13B.}
\label{table:architecture}
\end{table}

\subsection{Infrastructure}
Our Skywork-13B is trained on a cluster of 64 NVIDIA-HGX-A800 nodes, a total of 512 A800-80G SXM GPUs. Each node in the cluster is outfitted with high-speed 400GB/s NVLinks for intra-node communication and an 800Gb/s RoCE network for inter-node connectivity.
Our training framework is based on Megatron-LM \cite{megatronlm} library, designed to support the stable, prolonged training of large-scale models, accommodating thousands of GPUs and model sizes in the order of hundreds of billions parameters. 

Considering the relatively moderate size of our Skywork-13B model, we have avoided the use of GPU memory optimization techniques and parallel schemes that could impede speed. These include Tensor Model Parallelism \cite{megatronlm}, Sequence Parallelism \cite{sequence_parallel}, ZeRO-Stage2 \cite{zero}, and Checkpointing \cite{checkpointing}.
Instead, we have leveraged Data Parallelism (DP) with ZeRO-1 \cite{zero} and Pipeline Parallelism (PP) \cite{narayanan2021} as the primary parallelization strategies for training Skywork-13B. ZeRO-1 substantially diminishes the GPU memory footprint of the Adam optimizer state without increasing the burden on intercommunication. Pipeline Parallelism offers memory optimization at a minimal communication overhead, which decreases as the gradient accumulation step increases, thereby mitigating the slowdown of all-reduce as DP Size increases.
Regarding operator optimization, we adopted Flash Attention V2 \cite{flashattention, flashattention2}, a strategy that both optimizes GPU memory and expedites the training process.

Upon extensive preliminary experiments, we have decided to adopt the combination of  \texttt{DP256}, \texttt{PP2}, and ZeRO-1 as our distributed training strategy for Skywork-13B. With this configuration, we achieved a token throughput of 1873 per GPU per second and a model flops utilization (MFU) of 56.5\%. An overview of these experiments is provided in Appendix \ref{distributed_training}.
The training process of Skywork-13B spanned a total of 39 days.

\subsection{Training Details}
As outlined in Section \ref{two_stages}, the pre-training of Skywork-13B is executed in two stages: 
\begin{itemize}
    \item {\bf Stage-1:} General purpose pre-training on SkyPile-Main.
    \item {\bf Stage-2:} STEM-oriented continual pre-training on SkyPile-STEM.
\end{itemize}
In both stages, the model is trained using the standard auto-regressive language modeling objective, with context lengths fixed at 4096 tokens. The AdamW optimizer \cite{adamW}, applied for the training process, uses $\beta_1$ and $\beta_2$ values of 0.9 and 0.95, respectively. Throughout the pre-traning, we applied a weight decay of 0.1 and gradient clipping of 1.0. Our model was trained with bfloat16 mixed precision.

\begin{figure*}[ht]
\centering
\includegraphics[width=1.0\textwidth]{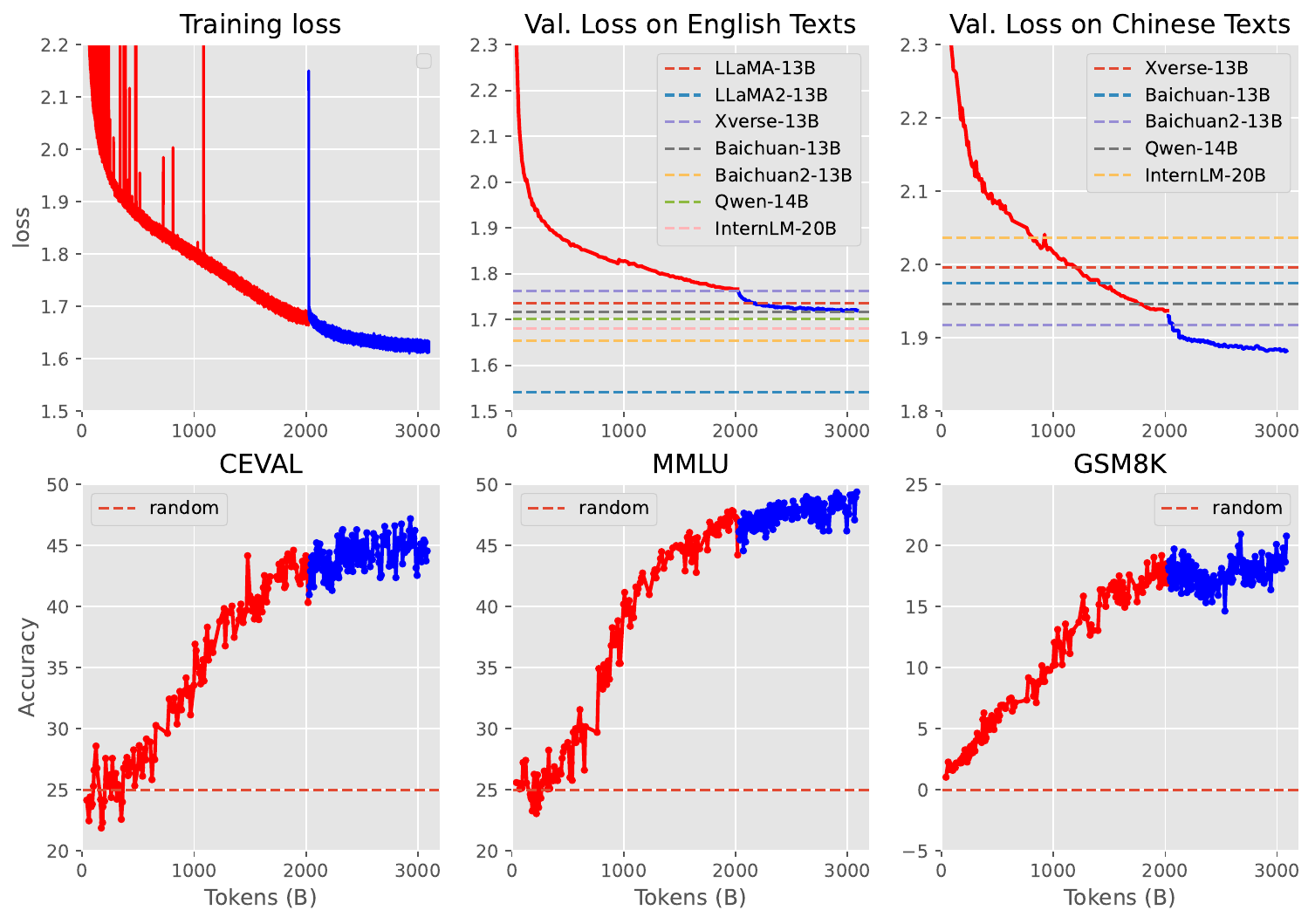}
\caption{Trajectory of important monitoring metrics during Stage-1 pre-training.
Top Left: Training loss.
Top Middle and Right: Validation loss on English and Chinese held-out sets of web texts.
The horizontal dashed lines in the middle and right plots correspond to the evaluated language modeling loss for several similar-sized open LLMs.
Bottom: Benchmark results on CEVAL, MMLU and GSM8K respectively.
Stage-1 pre-training consists of two sequential training sessions, represented by different colors in the loss curves (red for session $0\sim2$T and blue for session $2\sim3$T). 
}
\label{validation_loss}
\end{figure*}

\subsubsection{Stage-1 Pre-training}
In the first stage, our Skywork-13B model is trained from scratch on SkyPile-Main for over three trillion tokens. 
This stage consists of two sequential training sessions, covering the first $0\sim2$T  tokens and the subsequent $2\sim3$T tokens, respectively.

Our initial plan was to train Skywork-13B for two trillion tokens. We launched a training session accordingly, with a cosine learning rate schedule that gradually decays from a peak learning rate of 6e$-4$ to a final learning rate of 6e$-5$.
In Figure. \ref{validation_loss}, we report in red curves the evolution of language modeling losses and several benchmark results of our Skywork-13B during this session. 
It is evident that by the end of this session, the model had not reached saturation. We hypothesized that the model could further benefit from additional pre-training, prompting us to launch a secondary training session targeting an additional one trillion tokens.

\begin{figure}[h]
\centering
\includegraphics[width=0.45\textwidth]{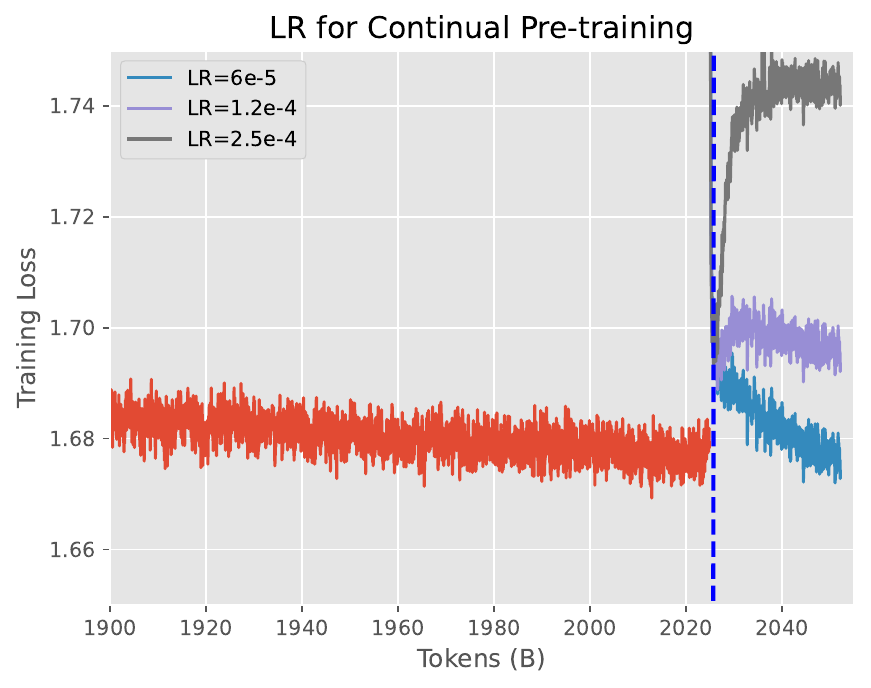} 
\caption{Test runs for tuning the learning rate of the $2\sim3$T training session. 
It can be seen that 6e-5, which is the terminal learning rate from $0\sim2$T training session, yields the best result.
}
\label{lr_contrain}
\end{figure}

The second training session utilized a slightly different composition of training data compared to the initial $0\sim2$T session, as data from certain sources had been depleted and fresh sources were introduced. Owing to the shift in the training distribution, we meticulously tuned the learning rate parameter, eventually deciding on a constant learning rate of 6e-5 for the $2\sim3$T session.
In Figure. \ref{lr_contrain}, we illustrate the model losses under varying learning rate conditions. Results indicate that a higher learning rate leads to escalations in training loss which we deem too costly to reverse. 
The impact of the second training session is depicted in blue curves of Fig. \ref{validation_loss}. The enhancement in the model's performance continues, albeit at a decelerating pace.
Interestingly, although our Skywork-13B trails in the realm of English language modeling, it significantly surpasses all other comparable open LLMs in Chinese language modeling.
In Section \ref{section:lm_test}, we will confirm that the superiority of our Skywork-13B in Chinese language modeling is not only true on our validation set, it also holds true on a number of test sets sourced from diverse domains. 

More results can be found in Appendix (see Figure \ref{fig:benchmark_trajectory}).

\subsubsection{Stage-2 Pre-training}
The primary aim of Stage-2 pre-training is to augment the model with capabilities pertinent to STEM disciplines. The data utilized in this stage comprises an approximate 20\% from SkyPile-STEM and 80\% from SkyPile-Main, amassing a total of roughly 130 billion tokens. A constant learning rate of 6e$-5$ is adopted, maintaining parity with the terminal learning rate used in Stage-1 pre-training

Consequent to the data distribution shift from Stage-1 to Stage-2, it becomes crucial to meticulously calibrate the sampling ratio between the different data sources. Initial experiments revealed that a gradual increment in the SkyPile-STEM ratio yielded the most effective results. Therefore, for the actual Stage-2 pre-training phase, we implemented a sampling plan that commenced with 10\% of SkyPile-STEM initially, gradually escalating to a peak of 40\% towards the conclusion of the training.

This training strategy proved successful in maintaining the stability of the model's language modeling validation loss while enabling an optimum transfer of STEM knowledge. The extended training period ensures a comprehensive assimilation of STEM-related knowledge into the model without causing significant disturbance to the pre-existing learned information.

\begin{figure}[h]
\centering
\includegraphics[width=0.45\textwidth]{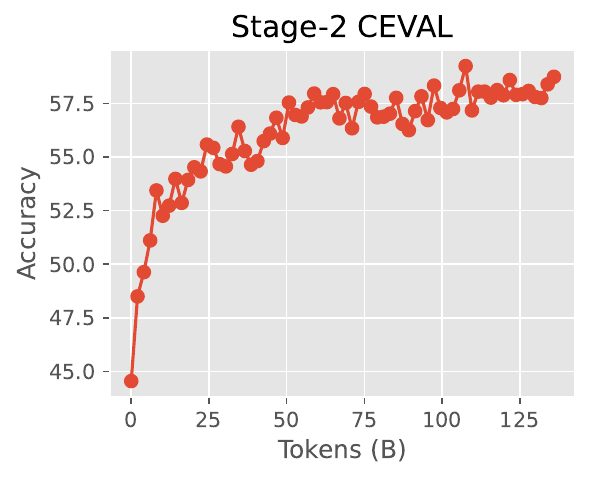} 
\caption{Evolution of CEVAL score during Stage-2 pre-training.
}
\label{fig:stage2_pretrain}
\end{figure}

The impact of Stage-2 pre-training is illustrated in Figure \ref{fig:stage2_pretrain}, which presents the progression of the CEVAL benchmark score. The evolution of scores on other STEM-related benchmarks, such as GSM8K, mirrors a similar trend. Improvements in individual subjects of the CEVAL can be found in Table \ref{tab:ceval_detail} (see appendix).

\section{Evaluation\label{evaluation}}
\subsection{Baselines}
We compare the performance of our Skywork-13B with open models that are similar in size, including LLaMA-13B \cite{llama}, LLaMA2-13B \cite{llama2}, Baichuan-13B, Baichuan2-13B \cite{baichuan}, Xverse-13B \cite{xverse13b}, IntermLM-20B \cite{internlm}. A summary of these models can be found in Table \ref{tab:model_tokens}.  
\begin{table}[t]
\centering
\resizebox{0.45\textwidth}{!}{%
\begin{tabular}{lcc}
\toprule
\textbf{Model} & \textbf{\#Tokens} & \textbf{Language}  \\
\midrule
OpenLLaMA-13B & 1.0T & English \\
LLaMA-13B  & 1.0T & English  \\
LLaMA2-13B  & 2.0T & English \\
Baichuan-13B & 1.4T & English \& Chinese  \\
Baichuan2-13B & 2.6T &  English \& Chinese   \\
Xverse-13B  & 1.4T &  English \& Chinese    \\
InternLM-20B & 2.3T &  English \& Chinese   \\
\midrule
Skywork-13B & \underline{3.2T} &  English \& Chinese    \\
\bottomrule
\end{tabular}
}
\caption{Details of various models. The column labeled "\#Tokens" indicates the quantity of training tokens used by each model, whereas the "Language" column specifies the primary languages supported by each model. }
\label{tab:model_tokens}
\end{table}


\subsection{Benchmark Evaluation}
We focus on the following popular benchmarks:
\begin{itemize}

\item MMLU \cite{mmlu}: MMLU is a benchmark designed to measure knowledge acquired during pre-training. The benchmark covers 57 subjects across STEM, the humanities, the social sciences, and more, ranging in difficulty from an elementary level to an advanced professional level. It tests both world knowledge and problem solving ability. 
\item CEVAL \cite{ceval} and CMMLU \cite{cmmlu}: Those are Chinese benchmarks that mimick MMLU. CEVAL consists of 13948 multi-choice questions spanning 52 diverse disciplines and four difficulty levels. CMMLU covers 67 disciplines that span from elementary to advanced professional levels. 
\item GSM8K \cite{gsm8k}: This dataset consists of 8500 high-quality grade school math word problems created by human writers. These multi-step problems require between 2 and 8 steps to solve. GSM8K is usually used in benchmarking multi-step mathematical reasoning ability of LLMs.
\end{itemize}

In Table \ref{tab:main_results} we present a comparison of performance results from different models on these benchmarks. The metrics for CEVAL, CMMLU and MMLU are 5-shot accuracy, while for GSM8K it is 8-shot accuracy. Higher numbers indicate better performance.
It can be seen that our Skywork-13B achieves the highest score on both the CEVAL and MMLU  and GSM8K benchmarks, with scores of 60.6 and 62.1 and 55.8 respectively. On the CMMLU benchmark, Baichuan2-13B achieves the highest performance with a score of 62.0. 

In summary, our Skywork model has demonstrated exceptional performance across a diverse range of comprehensive benchmark tests.

Results of individual subjects of the CEVAL can be found in Table \ref{tab:ceval_detail}. Results of other benchmarks can be found in Appendix \ref{more_results}.

\begin{table*}[ht]
\centering
\begin{tabular}{lcccc}
\toprule
\textbf{Model} & \textbf{CEVAL} & \textbf{CMMLU} & \textbf{MMLU} & \textbf{GSM8K} \\
\midrule
OpenLLaMA-13B & 27.1 & 26.7 & 42.7 & 12.4 \\
LLaMA-13B & 35.5 & 31.2 & 46.9 & 17.8 \\
LLaMA-2-13B & 36.5 & 36.6 & 54.8 & 28.7 \\
Baichuan-13B & 52.4 & 55.3 & 51.6 & 26.6 \\
Baichuan2-13B & 58.1 & \underline{62.0} & 59.2 & 52.8 \\
XVERSE-13B & 54.7 & - & 55.1 & - \\
InternLM-20B & 58.8 & - & 62.0 & 52.6 \\
\midrule
Skywork-13B & \underline{60.6} & 61.8 & \underline{62.1} & \underline{55.8} \\
\bottomrule
\end{tabular}
\caption{Comparison of results on popular benchmarks. Best result in each column is underlined.
It can be seen that our Skywork-13B consistently perform well across the different benchmarks, indicating its overall robustness.}
\label{tab:main_results}
\end{table*}

\subsection{Language Modeling Results \label{section:lm_test}}
\subsubsection{LM as a solution to benchmark overfitting}
Conventional benchmarks for evaluating LLMs often rely on static datasets of human-annotated examples. A core issue with this approach is that updating the test samples regularly is difficult and costly. Over time, the static test sets tend to be overfitted, producing misleading benchmark results.

We propose language modeling evaluations as a compelling alternative. Perplexity in language modeling acts as a proxy metric strongly linked to performance on diverse downstream tasks (see Figure \ref{fig:loss_vs_metric}). Since language modeling solely requires unlabeled natural text, it eliminates the need for expensive human annotation. Constructing and revising language modeling test sets is low-cost, as new data can be readily sampled from newly published content. Additionally, if a test set becomes compromised, fresh test data can quickly be sampled as a replacement.

\subsubsection{Construction of diverse LM testsets}
We compare the language modeling capabilities of various language models with our Skywork-13B, focusing on Chinese language. 

To conduct a robust evaluation of language modeling capability, we have separately collected a diverse corpus of texts from a myriad of websites, each labeled according to its respective domain. The domains we cover span a wide spectrum, encompassing areas such as technology, movies, finance, to name a few.  These domain-specific evaluation datasets have also been open-sourced for public access\footnote{Github: \url{https://github.com/SkyworkAI/Skywork/tree/main/data/eval_loss}}.

We ensure that every test sample consists of documents or user posts published \emph{after} September 1, 2023. This cut-off date guarantees that no test sample was inadvertently included during the pre-training of any evaluated language model. Specifically, SkyPile's cut-off date is June 30, 2023, and the majority of models under evaluation were released prior to August 31.

Note that while the held-out validation set used to monitor the training progress (as shown in Figure \ref{validation_loss}) of our model can also serve this purpose, it has the same distribution (web texts) as the bulk of the training corpus, thus may lead to overly optimistic estimate of the actual language modeling capability of the model.
More details on the sources of the test samples and the underlying data collection pipeline can be found in Appendix \ref{lm_test}.

\subsubsection{Results}
The results of our language modeling evaluation are presented in Table \ref{table:zh_test_ppl}, where results from ChatGLM3-6B \cite{chatglm3}, MOSS-7B \cite{moss}, Baichuan2-7B \cite{baichuan}, Qwen-7B \cite{qwen}, InternLM-7B \cite{internlm} and Aquilla2-34B are also included.

It can be seen that our Skywork-13B model shows the best performance overall, obtaining the lowest average perplexity score of 9.42. It also exhibits the best performance across individual domains, achieving the lowest perplexity scores in tech (11.58), movie (21.84), government (4.76), and finance (4.92) domains. It excels not only in surpassing the performance of models of a similar size, but also in outperforming significantly larger models such as InternLM-20B and Aquila2-34B.

We attribute the excellent language modeling performance of our Skywork-13B to the quality of our training corpus. Details on rigorous data filtering pipeline are described in Section \ref{skypile}.

\begin{table*}[ht]
\centering
\begin{tabular}{l|cccccc|c}
\toprule
 & \textbf{Tech} & \textbf{Movie} & \textbf{Gov.} & \textbf{Game} & \textbf{Finance} & \textbf{General} & \textbf{Average} \\
 \midrule
{ChatGLM3-6B} & 12.48 & 23.48 & 5.07 & 18.45 & 5.67 & 7.47 & 10.25 \\
{MOSS-7B} & 20.83 & 39.66 & 11.08 & 31.24 & 10.59 & 13.25 & 18.50 \\
{InternLM-7B} & 13.43 & 24.9 & 5.88 & 19.78 & 6.17 & 8.10 & 11.17 \\
{Qwen-7B} & 13.39 & 25.16 & 5.55 & 19.26 & 5.76 & 7.78 & 10.83 \\
{Baichuan2-7B} & 12.89 & 23.26 & 5.34 & 18.36 & 5.68 & 7.62 & 10.41 \\
\midrule
LLaMA2-13B & 23.26 & 50.66 & 18.09 & 32.52 & 14.85 & 16.55 & 23.54 \\
Xverse-13B & 12.55 & 23.49 & 5.20 & 17.69 & 5.54 & 7.46 & 10.19 \\
Baichuan-13B & 12.38 & 22.46 & 5.21 & 17.59 & 5.42 & 7.37 & 10.03 \\
Baichuan2-13B & 12.14 & 21.85 & 5.05 & 17.15 & 5.35 & 7.24 & 9.81 \\
Qwen-14B & 11.90 & 22.43 & 4.89 & \underline{16.94} & 5.24 & 7.03 & 9.67 \\
InternLM-20B & 12.34 & 22.06 & 5.75 & 17.45 & 5.73 & 7.78 & 10.34 \\
Aquila2-34B & 14.62 & 29.09 & 5.72 & 21.78 & 5.83 & 8.45 & 11.73 \\
\midrule
Skywork-13B & \underline{11.58} & \underline{21.84} & \underline{4.76} & 17.28 & \underline{4.92} & \underline{6.82} & \underline{9.42} \\
\bottomrule
\end{tabular}
\caption{Comparative analysis of language modeling capabilities across diverse domains. Performance is measured using perplexity (lower values is better). Underlined figures correspond to the best result in each column.}
\label{table:zh_test_ppl}
\end{table*}

\section{Discussion}
In this section, we delve into the benefits and associated risks of pre-training on the in-domain data\footnote{The term ``in-domain data'' is a vague one that refers to any data with distribution closely resembling to that of the task data. For instance, the training data of a task is trivially in-domain data for that task. GPT-4 generated data with few-shot task examples can also be considered as in-domain data for that task. 
} of benchmark tasks.

\subsection{Effect of pre-training on in-domain data}
Pre-trained language models, or foundation models, are intended to be used in transfer learning as a general purpose backbone. As a foundation model in itself has little usage other than sentence completion, the quality of a foundation model is typically evaluated in terms of its performance in those tasks. Apparently, when it comes to improve a foundation model's quality as measured by its task performance, it is always far more efficient to train the model on in-domain data of that task \cite{hernandez2021scaling, chung2022scaling} , as compared to general-purpose data (web texts). 

We have shown that Stage-2 pre-training significantly amplifies our Skywork-13B's STEM related capabilities, leading to a substantial improvement in performance on STEM-related tasks. Now we show that it is even possible to enhance a much weaker base model, i.e., an intermediate checkpoint, using only a fraction of the data and compute used in Stage-2 pre-training. 

\begin{table}[ht]
\renewcommand{\arraystretch}{1.0} 
\centering
\resizebox{0.45\textwidth}{!}{%
\begin{tabular}{l|cc|cc}
\toprule
 & CEVAL & GSM8K & En Loss & Zh Loss \\
\midrule
Before & 28.3 & 6.9 & 1.86 & 2.08\\
After  & 50.8 & 40.7 & 2.09 & 2.21\\
\midrule
$\Delta$ & +22.5 & +33.8 & +0.23 & +0.13 \\
\bottomrule
\end{tabular}
}
\caption{The impact of pre-training on a 0.5T checkpoint of Skywork-13B using only 1B tokens. The training data is sourced from a subset of our SkyPile-STEM corpus.
The columns ``En Loss'' and ``Zh Loss'' show the model's validation loss on held-out sets of English and Chinese web texts, respectively. 
}
\label{table:in_domain}
\end{table}

Table \ref{table:in_domain} presents the CEVAL and GSM8K scores before and after pre-training on in-domain data, utilizing a relatively weak model checkpoint that has only undergone 0.5T pre-training. The results indicate that after pre-training with merely 1B tokens of in-domain data, a weak model, initially performing only slightly better than random at CEVAL and GSM8K, can surpass the performance of our strongest Skywork-13B (3T) backbone without in-domain pre-training. However, this comes at the cost of significant degradation in language modeling performance, as evidenced by the higher loss on both tasks, shown in the two rightmost columns of the table.

\subsection{Pre-training on in-domain data: a common practice?}
It is of interest to explore whether popular foundational models are pre-trained on in-domain data. In pursuit of this, we delve into the GSM8K datasets, equipped with official train/test splits and comprehensive solutions. We evaluate an LLM's language modeling loss on three datasets drawn from the same distribution:
1) The official GSM8K training set, 
2) The official GSM8K test set, 
3) A set composed of GSM8K-like samples generated by GPT-4. 
The corresponding losses are denoted as $L_{train}$, $L_{test}$, and $L_{ref}$, respectively. Theoretically, if a language model has not been exposed to any of the three datasets during pre-training, the three losses $L_{train}$, $L_{test}$, and $L_{ref}$ should be approximately equivalent. However, if the model has been pre-trained on the training set or if the test data has been inadvertently exposed during the pre-training process, we would anticipate a notable discrepancy between $L_{train}$, $L_{test}$, and $L_{ref}$.

Our results are outlined in Table \ref{in-domain-pre-training}, which also reports the differences in losses $\Delta_1=L_{test}-L_{ref}$ and $\Delta_2=L_{test} - L_{train}$. Notably, the $\Delta_2$ column reveals that for most models, the language modeling loss on the GSM8K training and test splits are almost identical. However, models such as ChatGLM3-6B, Baichuan2-13B, Qwen-7B/14B, and Aquila2-34B display markedly lower loss on the training split than on the test split. Consequently, we postulate that these models may have been considerably pre-trained on GSM8K training split or similar data.


Moreover, we notice one particular anomaly in the $\Delta_1$ column, indicating the significantly lower $L_{test}$ loss compared to $L_{ref}$, which is interesting to further study for better understanding.

\begin{table*}[ht]
\centering
\begin{tabular}{l|c|c|c|c|c}
\toprule
 & $L_{test}$ & $L_{train}$ & $L_{ref}$  & $\Delta_1$ & $\Delta_2$ \\
\midrule
ChatGLM3-6B  & 0.99 & 0.78 & 0.99 & 0.0 & \colorbox{gray!30}{0.21}\\
MOSS-7B      & 1.51 & 1.52 & 1.49     & 0.02 & $-0.01$ \\
InternLM-7B  & 1.21 & 1.12 & 1.27     & -0.06 & 0.09 \\
Qwen-7B      & 1.07 & 0.64 & 1.10     & -0.03 & \colorbox{gray!30}{0.43} \\
Baichuan2-7B & 1.41 & 1.42 & 1.36     & 0.05 & $-0.01$\\
\midrule
LLaMA-13B & 1.41 & 1.42 & 1.36 & 0.05 & $-0.01$ \\
LLaMA2-13B  & 1.36 & 1.38 & 1.33 & 0.03 &  $-0.01$ \\
Xverse-13B & 1.42 & 1.43 & 1.39 & 0.03 & $-0.01$\\
Baichuan-13B & 1.41 & 1.42 & 1.37 & 0.04 & $-0.01$\\
Baichuan2-13B & 1.09 & 0.72 & 1.12 & -0.03 & \colorbox{gray!30}{0.37} \\
Qwen-14B & 1.03 & {0.42} & 1.14 & -0.11 & \colorbox{gray!30}{0.61} \\
InternLM-20B & 1.20 & 1.09 & 1.19 & 0.01 & 0.11\\
Aquila2-34B & {0.78} & {0.39} & 1.29 & \colorbox{gray!30}{$-0.51$} & \colorbox{gray!30}{0.39}\\
\midrule
Skywork-13B & 1.01 & 0.97 & 1.00  & 0.01 & 0.04\\
\bottomrule
\end{tabular}
\caption{We evaluate the language modeling (LM) loss on samples (a sample is a concatenation of question and answer) from GSM8K dataset for several foundation models.
For each LLM, we compare LM loss on the training split ($L_{train}$), the test split ($L_{test}$), and a specially curated reference set ($L_{ref}$), generated by GPT-4, designed to mimic the GSM8K dataset. We also reports two key metrics: $\Delta_1=L_{test}-L_{ref}$, serving as an indicator of potential test data leakage during the training of the LLM, i.e., a lower value suggests possible leakage; and $\Delta_2=L_{test} - L_{train}$, which measures the degree of overfitting on the training split of the dataset. A higher value of $\Delta_2$ implies excessive overfitting. Outliers for both $\Delta_1$ and $\Delta_2$ are highlighted in gray.}
\label{in-domain-pre-training}
\end{table*}

\subsection{Pre-Training or Supervised Fine-Tuning?}
In the era preceding the advent of LLMs such as GPT-4 \cite{gpt4_sparks, gpt4_report} and Claude \cite{claude1}, supervised data for NLP tasks was generally scarce. This was because the process of data collection and annotation was both time-consuming and costly. Due to the scarcity of supervised data, NLP researchers rely on unsupervised pre-training techniques \cite{word2vec, elmo, GPT1, BERT} to improve downstream task performance via transfer learning, where supervised data is to be used only in the fine-tuning stage.
In this context, pre-training on in-domain (supervised) data was pointless, as it would defeat the purpose of pre-training itself (transfer learning). 

This reality has significantly shifted, however, with the emergence of powerful LLMs. This is because procuring large amounts of high quality supervised/in-domain data is now as simple as making a few API requests to these LLMs, and it is comparatively low-cost \cite{selfinstruct, alpaca}. This new reality blurs the boundary between pre-training and supervised fine-tuning, making it feasible to incorporate substantial amounts of supervised data into the pre-training phase \cite{phi1, phi2}.  After all, curated in-domain data, whether written by human annotators  or generated by LLM, are all form of human knowledge, and there is good reason for this knowledge to be absorbed into a foundation model.

That said, we believe that there is valid risk on the practice of targeted pre-training, in that it compromise fairness in benchmarking.  While through pre-training on in-domain data a model may excel at specific tasks, it remains uncertain how well it would perform on unseen tasks. Its capabilities may be overestimated based on the benchmark alone, which can lead to unfair comparisons between models and mislead users or stakeholders about the true capabilities of the model.

\section{Limitation}
Our pre-training approach for Skywork-13B involved a two-stage process: general purpose pre-training followed by domain-specific enhancement pre-training. However, it remains unclear whether this methodology can produce a model on par with, or superior to, a model trained in one stage on a mixed corpus. Further investigation is needed to determine the comparative effectiveness of these pre-training approaches.

Additionally, we have proposed using language modeling loss or perplexity as proxy metrics for monitoring and evaluating large language models. A limitation is that language modeling evaluation relies on the specific distribution used to sample test data, of which there are infinite possibilities. While language modeling perplexity over a given data distribution may predict performance on some tasks, it may not translate to other tasks. The correlation between language modeling and downstream performance could vary across different distributions and tasks.

\section{Conclusion}
Our work on Skywork-13B represents a significant leap forward in the development of open large language models. We believe that our comprehensive and transparent approach to the model's development will be a valuable resource for researchers in the field, fostering collaboration and open-source principles. Our two-stage training methodology, leveraging a segmented corpus, offers a novel approach for enhancing model capability in specific domain, while our method of monitoring the training progress provides a practical solution to the challenges of tracking the improvement of these models over time.

However, our work is more than just the creation of a new LLM. It is a call to action for the broader NLP community, urging a return to the principles of fairness, transparency, and the sharing of ideas that have historically fueled progress in the field. We hope that Skywork-13B will not only serve as a powerful tool for a wide range of applications but also inspire a renewed commitment to openness and cooperation in the development of future models.

\bibliographystyle{acl_natbib}
\bibliography{llm}
\appendix
\section{Details on GPT-7B vs. LLaMA-7B Experiment\label{GPT_vs_LLAMA}}
In a preliminary experiment, we compared the language modeling performance between GPT and LLaMA architecture in a controlled environment. We trained a 7B model with GPT architecture and a comparable 7B model with LLaMA architecture for 200B tokens sampled from the same corpus and with the same training parameters. Details are given in Table \ref{tab:detail_gpt_llama}.

\begin{table*}[h!]
\centering
\begin{tabular}{l|cc}
\toprule
& \textbf{GPT-7B} & \textbf{LLaMA-7B} \\ 
\midrule
Positional Embedding & Absolute & {Rotary} \\
Max Position Embeddings & 4096 & 4096 \\
Normalization & LayerNorm & {RMSNorm} \\
Activation & Gelu & {SwiGlu} \\ 
Attention & MHA & {MHA} \\
Num. Layers & 32 & 32 \\
Hidden Size & 4096 & 4096 \\
Num. Heads & 32 & 32 \\
FFN Size & 16384 & {11008} \\
Context Size & 4096 & 4096 \\
\midrule
Global Batch Size & 1024 & 1024 \\
Adam $\beta_1$ & 0.95 & 0.95 \\
Adam $\beta_2$ & 0.9 & 0.9 \\
Adam $\epsilon$ & 1.00e-8 & 1.00-8 \\
Precision & bf16 & bf16 \\
Peak Learning Rate & 3e-4 & 3e-4 \\
Min Learning Rate & 3e-5 & 3e-5 \\
Learning Rate Decay Steps & 43945 & 43945 \\
Learning Rate Decay Style & Cosine & Cosine \\
Warm-up Steps & 2000 steps & 2000 steps \\
Weight Decay & 0.1 & 0.1 \\
Dropout Probability & 0.1 & 0 \\
Gradient Clip & 1 & 1 \\
Total Steps & 51200 & 51200 \\
\bottomrule
\end{tabular}
\caption{Comparison of GPT-7B and LLaMA-7B. All variables are controlled in our experiment except for the differences in architecture.}
\label{tab:detail_gpt_llama}
\end{table*}

\section{Preliminary Experiments on Distributed Training \label{distributed_training}}
In Table \ref{tab:distributed_training_config} we report preliminary results obtained with various distributed training configurations on LLaMA-13B and Skywork-13B model architecture.
In both cases, the best throughput is achieved with \texttt{DP256} and \texttt{PP2} with ZERO-1 setting.

\begin{table*}[h!]
\begin{center}
\begin{tabular}{clccccc}
\toprule
Model & Strategy & Throughput & MFU & TFlops & Memory \\
\midrule
LLaMA2 & DP512 & - & - & - & OOM \\
LLaMA2 & DP256+PP2 & \underline{2045} & \underline{58.5} & \underline{182.6} & \underline{70.7} \\
LLaMA2 & DP256+TP2 & 1928 & 55.2 & 172.2 & 65.5 \\
LLaMA2 & DP128+TP2+PP2 & 1936 & 55.4 &  172.9 & 39.4 \\
LLaMA2 & DP128+PP4 & 1964 & 56.2 &  175.4 & 53.4 \\
LLaMA2 & DP128+TP4 & 1744 & 44.4 &  138.5 & 35.4 \\
\midrule
Skywork & DP512 & - & - & - & OOM \\
Skywork & DP256+PP2 & \underline{1873} & \underline{56.5} & \underline{176.2} & \underline{77.1} \\
Skywork & DP256+TP2 & 1775 & 53.5 & 167.0 & 67.9 \\
Skywork & DP128+TP2+PP2 & 1776 & 53.5 & 167.0 & 42.5 \\
Skywork & DP128+PP4 & 1828 & 55.1 &  171.9 & 58.7 \\
Skywork & DP128+TP4 & 1417 & 43.1 & 134.6 & 36.6 \\
\bottomrule
\end{tabular}
\caption{Compute effeciency achieved with different distributed training configurations.  We tested both LLaMA2-13B and Skywork-13B. Throughout the experiments, we use a global batch size of 4096 and a micro batch size of 1. When Tensor Parallelism is enabled, Sequence Parallelism is enabled as well. Throughput is measured in tokens processed per GPU per second, while Model Flops Utilization (MFU) is expressed as a percentage (\%). Memory usage is reported in Gigabytes (GB).}
\label{tab:distributed_training_config}
\end{center}
\end{table*}

\begin{table*}[ht]
\centering
\resizebox{0.95\textwidth}{!}{%
\begin{tabular}{lcccccccc}
\toprule
Models & BoolQ & PIQA & Winogrande & TriviaQA & RACE &  Hellaswag & ARC-E & ARC-C \\
\midrule
OpenLLaMA-13B & 77.6 & 79.5 & 72.0 & 60.2 & 42.4 & 76.0 & 78.9 & 48.6 \\
LLaMA-13B & 80.7 & 81.0 & \underline{76.2} & 65.0 & 43.4  & 80.1 & 82.1 & 54.7 \\
LLaMA2-13B & \underline{83.3} & \underline{81.7} & 75.8 & \underline{68.2} & 43.9  & \underline{81.5} & \underline{83.7} & \underline{57.0} \\
Baichuan-13B & 78.8 & 77.2 & 70.4 & 51.6 & 35.8 & 74.2 & 77.2 & 48.4 \\
Baichuan2-13B & 80.3 & 79.3 & 72.1 & 58.0 & 25.2  & 76.4 & 81.1 & 53.2 \\
Xverse-13B & 79.8 & 80.0 & 71.1 & 53.3 & 43.2 & 77.2 & 78.5 & 49.1  \\
\midrule
Skywork-13B & 82.9 & 79.9 & 72.2 & 54.0 & \underline{45.2} & 77.4 & 78.5 & 50.2 \\
\bottomrule
\end{tabular}
}
\caption{More English benchmarks results. As all of these models are more or less sensitive to the prompt template or number of shots, the reported results, which are reproduced by us, may be different to those from other sources.}
\label{table:en_benchmark}
\end{table*}

\section{More Benchmark Results\label{more_results}}
We also provide results of the following benchmarks in Table \ref{table:en_benchmark}:
\begin{itemize}
    \item TriviaQA \cite{triviaqa}: TriviaQA is a realistic text-based question answering dataset which includes 950K question-answer pairs from 662K documents collected from Wikipedia and the web.
    \item HellaSwag \cite{hellaswag}: HellaSWAG is a dataset that focuses on grounded commonsense inference.
    \item Winogrande \cite{winogrande}: WinoGrande is a dataset that focuses on commonsense reasoning.
    \item BoolQ \cite{boolq} BoolQ  is a question answering dataset for yes/no questions.
    \item PIQA \cite{piqa}: PIQA is a dataset for commonsense reasoning, and was created to investigate the physical knowledge of existing models in NLP.
    \item ARC \cite{arc}: ARC is a dataset consisting of multiple-choice question-answering tasks that focus on commonsense reasoning.
    \item RACE \cite{race} RACE is a dataset that focuses on reading comprehension.
\end{itemize}

\section{Details on LM Test Sets \label{lm_test}}
We established a daily crawl of published articles and user posts from a selection of widely used Chinese websites. This data collection process is distinct from the pipeline utilized to construct SkyPile. The purpose of gathering this data is to create independent language modeling test sets, categorized by their domain, for the evaluation of current open Language Learning Models (LLMs).

Below we describe the sources of these domain testsets:

\begin{itemize}
\item {\bf Technology:} AI related articles from (\url{36kr.com}). This website provides timely and comprehensive news articles about startups, technology, and business trends, primarily in the Chinese market. 
\item {\bf Movie:} User written movie reviews from Douban (\url{douban.com}). Douban is a popular social networking service in China that offers a platform for users to share their opinions and create content related to movies, books, and music. It is one of the most influential web 2.0 websites in China and has a strong focus on user-generated content.
\item {\bf Government:} News from website of People's Daily (\url{www.people.com.cn}), which is  the most influential and authoritative newspapers in China. The language used in the news is typically formal Standard Mandarin and carries an authoritative tone. 
\item {\bf Game:} Articles from Gcores (\url{www.gcores.com}). This is a Chinese digital media platform dedicated to video games, tech trends, and geek culture. The platform features a wide range of original content, including news articles, podcast episodes, videos, and independent games.
\item {\bf Finance:} News from finance section of Sina (\url{finance.sina.com.cn}). It is one of China's leading online media companies, offers a comprehensive suite of financial information and services. It covers a broad range of topics including stock markets, forex, commodities, real estate, and personal finance.
\item {\bf General:} News from Jiemian News (\url{www.jiemian.com}). Jiemian is a prominent Chinese digital media platform known for its in-depth and high-quality journalism. It covers a wide range of topics, including politics, economy, culture, technology, finance, and lifestyle.
\end{itemize}

\begin{table*}[ht]
\centering
\resizebox{0.7\textwidth}{!}{%
\begin{tabular}{|l|ccc|}
\toprule
\textbf{Subject} & \textbf{Stage-1} & \textbf{Stage-2} & \textbf{Boost} \\
\midrule
Accountant & 40.8 & 49.0 & 8.2 \\
Advanced Mathematics & 26.3 & 42.1 & 15.8 \\
Art Studies & 60.6 & 72.7 & 12.1 \\
Basic Medicine & 42.1 & 57.9 & 15.8 \\
Business Administration & 42.4 & 48.5 & 6.1 \\
Chinese Language and Literature & 47.8 & 56.5 & 8.7 \\
Civil Servant & 40.4 & 66.0 & 25.5 \\
Clinical Medicine & 36.4 & 40.9 & 4.5 \\
College Chemistry & 37.5 & 50.0 & 12.5 \\
College Economics & 52.7 & 47.3 & -5.5 \\
College Physics & 15.8 & 36.8 & 21.1 \\
College Programming & 51.4 & 51.4 & 0.0 \\
Computer Architecture & 33.3 & 52.4 & 19.0 \\
Computer Network & 21.1 & 26.3 & 5.3 \\
Discrete Mathematics & 50.0 & 18.8 & -31.3 \\
Education Science & 44.8 & 75.9 & 31.0 \\
Electrical Engineer & 35.1 & 35.1 & 0.0 \\
Environmental Impact Assessment Engineer & 45.2 & 51.6 & 6.5 \\
Fire Engineer & 45.2 & 51.6 & 6.5 \\
High School Biology & 42.1 & 78.9 & 36.8 \\
High School Chemistry & 36.8 & 63.2 & 26.3 \\
High School Chinese & 26.3 & 42.1 & 15.8 \\
High School Geography & 36.8 & 78.9 & 42.1 \\
High School History & 80.0 & 80.0 & 0.0 \\
High School Mathematics & 27.8 & 16.7 & -11.1 \\
High School Physics & 42.1 & 57.9 & 15.8 \\
High School Politics & 47.4 & 84.2 & 36.8 \\
Ideological and Moral Cultivation & 84.2 & 100.0 & 15.8 \\
Law & 33.3 & 45.8 & 12.5 \\
Legal Professional & 39.1 & 52.2 & 13.0 \\
Logic & 50.0 & 45.5 & -4.5 \\
Mao Zedong Thought & 70.8 & 83.3 & 12.5 \\
Marxism & 57.9 & 63.2 & 5.3 \\
Metrology Engineer & 37.5 & 58.3 & 20.8 \\
Middle School Biology & 76.2 & 95.2 & 19.0 \\
Middle School Chemistry & 30.0 & 95.0 & 65.0 \\
Middle School Geography & 41.7 & 83.3 & 41.7 \\
Middle School History & 59.1 & 81.8 & 22.7 \\
Middle School Mathematics & 15.8 & 36.8 & 21.1 \\
Middle School Physics & 42.1 & 73.7 & 31.6 \\
Middle School Politics & 52.4 & 90.5 & 38.1 \\
Modern Chinese History & 47.8 & 73.9 & 26.1 \\
Operating System & 52.6 & 47.4 & -5.3 \\
Physician & 46.9 & 57.1 & 10.2 \\
Plant Protection & 63.6 & 63.6 & 0.0 \\
Probability and Statistics & 27.8 & 33.3 & 5.6 \\
Professional Tour Guide & 69.0 & 65.5 & -3.4 \\
Sports Science & 42.1 & 52.6 & 10.5 \\
Tax Accountant & 30.6 & 49.0 & 18.4 \\
Teacher Qualification & 61.4 & 84.1 & 22.7 \\
Urban and Rural Planner & 50 & 67.4 & 17.4 \\
Veterinary Medicine	& 26.1	& 60.9	& 34.8 \\
\bottomrule
\end{tabular}
}
\caption{Details on CEVAL benchmark results.}
\label{tab:ceval_detail}
\end{table*}

\begin{figure*}[ht]
\centering
\includegraphics[width=0.75\textwidth]{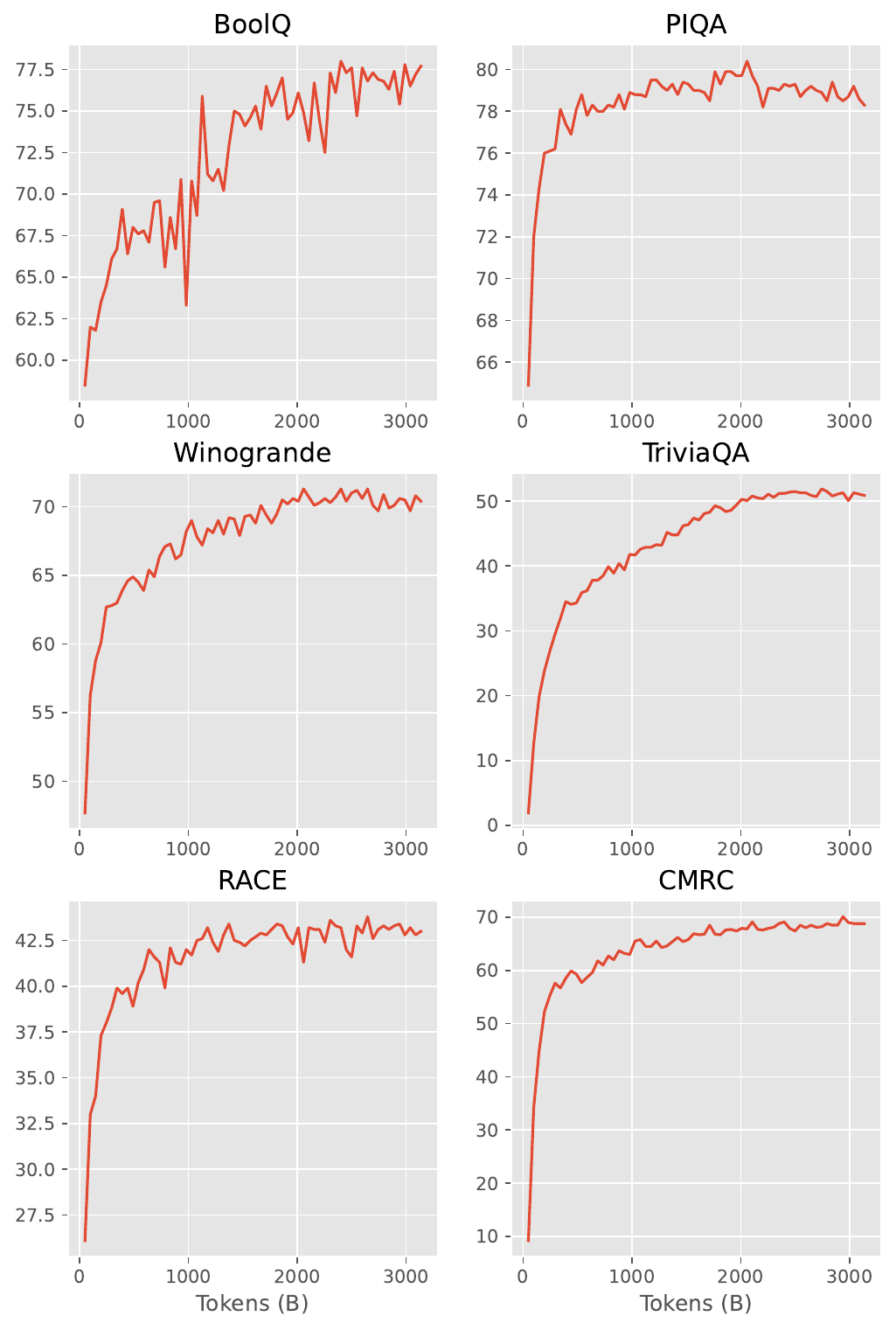}
\caption{Performance of the Skywork-13B on various benchmarks during Stage-1 pre-training. Benchmarks include BoolQ, PIQA, Winogrande, TriviaQA, RACE, and CMRC.}
\label{fig:benchmark_trajectory}
\end{figure*}

\end{document}